\documentclass[10pt,conference,a4paper]{IEEEtran}

\usepackage{hyperref}
\hypersetup{
    colorlinks=true,
    linkcolor=,
}

\usepackage{cite}

\ifCLASSINFOpdf
   \usepackage[pdftex]{graphicx}
\else
   \usepackage[dvips]{graphicx}
\fi
\usepackage{amssymb}
\usepackage{amsmath}
\ifCLASSOPTIONcompsoc
 \usepackage[caption=false,font=normalsize,labelfont=sf,textfont=sf]{subfig}
\else
 \usepackage[caption=false,font=footnotesize]{subfig}
\fi
\usepackage{multirow}
\usepackage{xcolor}
\usepackage{microtype}
\usepackage{tikz}
\usetikzlibrary{positioning,calc,spy}
\def\etal{\emph{et al}.}
\usepackage[T1]{fontenc}

\newcommand\nnfootnote[1]{%
  \begin{NoHyper}
  \renewcommand\thefootnote{}\footnote{#1}%
  \addtocounter{footnote}{-1}%
  \end{NoHyper}
}

\begin{document}
%
\title{ Mutual Information based Method for Unsupervised Disentanglement of Video Representation}

\author{\IEEEauthorblockN{P Aditya Sreekar\textsuperscript{*},
Ujjwal Tiwari\textsuperscript{*}, Anoop Namboodiri}
\IEEEauthorblockA{Center for Visual Information Technology\\
International Institute of Information Technology, Hyderabad\\
Email: paditya.sreekar@research.iiit.ac.in,
ujjwal.t@research.iiit.ac.in,
anoop@iiit.ac.in}}

\maketitle
\nnfootnote{\textsuperscript{*}Equal contribution}
\begin{abstract}
Video Prediction is a challenging but interesting task of predicting future frames from a given set context frames that belong to a video sequence. Video prediction models have  prospective applications in maneuver planning, healthcare, autonomous navigation and simulation. One of the major challenges in future frame generation is the high dimensional nature of visual data. To handle this, we propose a Mutual Information Predictive Auto-Encoder (MIPAE) framework that reduces the task of predicting high dimensional video frames by factorising video representations into content and low dimensional pose latent variables. Our approach leverages the temporal structure in the latent generative factors of video sequences by applying a novel mutual information loss to learn disentangled video representations. A standard LSTM network is used to predict these low dimensional pose representations. Content and the predicted pose representations are decoded to generate future frames. We also propose a metric based on mutual information gap (MIG) to quantitatively assess the effectiveness of disentanglement on DSprites and MPI3D-real datasets. MIG scores corroborate the visual superiority of frames predicted by MIPAE. We also compare our method quantitatively on LPIPS, SSIM and PSNR evaluation metrics.

\end{abstract}


%
\IEEEpeerreviewmaketitle

\section{Introduction}
Humans through experience learn to predict possible future states of the visual world by conditioning on the past. This key foresight has helped us in planning our actions ahead of time in highly dynamic real-world settings \cite{bubic2010}. In this work, our goal is to design an  intelligent system that is capable of predicting future frames of a video sequence. That is, given a sub-sequence of context frames, a generative model is trained to predict
plausible future frames. This problem of video frame prediction has been studied in variety of different contexts such as anomaly detection \cite{Liu_2018_CVPR, medel2016anomaly}, robotics \cite{finn2016unsupervised,finn2017deep}, and healthcare \cite{paxton2019visual}. Recent advancements in predictive auto-encoder based generative modelling methods such as the advent of GANs\cite{goodfellow2014generative} and VAE\cite{kingma2013auto} have substantially improved the visual quality of the predicted frames. However, modelling the predictive distribution of future frames is challenging due to the high-dimentional nature of video frames. 

Concurrent video prediction methods overcome the challenge of making predictions in high dimensional pixel space by factorising video representations into a low dimensional temporally varying component and another temporally consistent component. Tulyakov \etal \cite{Tulyakov_2018_CVPR} factorised video representations into time dependent pose and time independent content representations. Similarly, Vondrick-Pirsiavash \etal\cite{Vondrick2016GeneratingVW} decomposed video into salient (foreground) and non-salient (background) regions. In this paper, we introduce a novel predictive auto-encoder, referred to as MIPAE, that factorises the latent space representation of a video into two components, time dependent pose and time independent content latent variables. Our interpretation of the factorised latent space is similar to that of DRNET \cite{denton2017unsupervised}.
\begin{figure}[t]
    \centering
    \begin{tabular}{c}
        \includegraphics[width=0.47\textwidth]{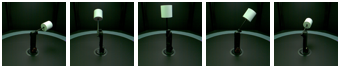}\\
        \includegraphics[width=0.47\textwidth]{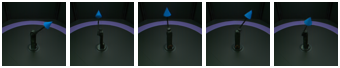}
    \end{tabular}
    \caption{pose/content disentanglement by MIPAE framework: Position latent variables form the target sequence(top) are combined with content representation of object from the second sequence. The object in the generated sequence (bottom) follows the pose description of the target sequence.}
    \label{fig:teaser_image}
\end{figure}

DRNET \cite{denton2017unsupervised} disentangled video representation into time dependent pose and time independent content latent representations. In their work, an adversarial loss is applied on the pose latent variables to achieve  pose/content disentanglement. The adversarial objective enforces the pose latent variables to be indiscriminate for different video sequences, ensuring that they do not encode any content information. In the proposed work, the application of adversarial loss on pose latent representations has been formalised as reducing mutual information between pose representations across time. This ensures that pose latent factors contain minimum mutual information, thus enforcing that they should not contain any content information (Fig. \ref{fig:teaser_image}). We train a LSTM model conditioned on the content latent representation of the last observed frame to predict the low dimensional pose representation for future frames. The predicted pose and content representation is used to generate the next frame of the video sequence.  

Mutual Information (MI) is a fundamental measure in Information theory that quantifies the amount of dependency between two random variables. By virtue of the recent advancements in deep learning, a number of methods \cite{belghazi2018mine,lin2019data, pmlr-v97-poole19a} use Neural Estimators to approximate MI. The ability of these methods in approximating MI between two random variables, given samples from the joint and marginal distributions has inspired multiple applications of mutual information based loss term for learning disentangled data representations \cite{chen2018isolating,kim2018disentangling}, \cite{chen2016infogan}. In the proposed MIPAE framework, Jenson-Shannon lower bound estimate of MI \cite{pmlr-v97-poole19a} is penalised between pose latent representations across time for proper pose/content factorisation. The contributions of this paper are as follows:

\begin{itemize}
    \item To the best of our knowledge, this work is the first to propose and empirically validate that minimising mutual information between pose latent variables across time in predictive auto-encoder setting for the task of frame prediction leads to factorised latent space representation of videos.

    \item We adopt and present a metric based on Mutual Information Gap \cite{chen2018isolating} to quantitatively measure the effectiveness of disentanglement between the factorized latent variables - pose and content. This score has also been used for conclusive evaluation of our disentanglement method.
\end{itemize}

\section{Related Work}
\subsection{Video Prediction}

Previously reported methods predicted future frames by minimising reconstruction error in predictive auto-encoder or recurrent network setting \cite{srivastava2015unsupervised,ranzato2014video,xingjian2015convolutional}. These methods assume the process of video generation to be deterministic. This assumption leads to blurry frame predictions on real-world videos exhibiting scene with stochastic dynamics as the model predicts an average of the possible future frames. \cite{denton2018stochastic,babaeizadeh2017stochastic} used VAE-based latent variable models to account for the inherent stochasticity in real world video sequences. While, these networks can model distributions over possible futures, the predictive distribution is still fully factorized over pixels which tends to produce blurry
predictions. To alleviate this issues of blurry frame prediction, SAVP \cite{lee2018stochastic} used adversarial training in a VAE-GAN setting to produce sharp and stochastic predictions. The aim of our MIPAE framework is to not only obtain visually sharp frame predictions but also produce proper disentanglement of video representations that significantly reduces the complexity of visual frame prediction. 

Transformation based methods predict subsequent frames by transforming previous frame through a constrained geometric transformation \cite{finn2016unsupervised,vondrick2017generating}. These methods primarily focus on modeling motion rather than on reconstructing appearances by exploiting the temporal consistency in video sequences. Finn-Goodfellow \etal \cite{finn2016unsupervised} modeled motion of masked out groups of pixels in a Conv-LSTM framework by predicting transformation kernels at each time step in an action conditioned setting. Vondrick-Torrabla \etal \cite{vondrick2017generating} generated transformations for each pixel using adversarial training to constraint the set of plausible transformations. Learning non-linear transformations between consecutive frames in a video sequence that exhibit stochastic dynamics is extremely challenging due to the high-dimentionality of video frames. Complementary to these methods, our framework predicts a single low dimensional pose vector rather than predicting non-linear transformations between consecutive frames.

Disentangling representation for the task of video prediction ha been explored by \cite{villegas2017decomposing,Tulyakov_2018_CVPR,denton2017unsupervised,hsieh2018learning}. MCNet \cite{villegas2017decomposing} disentangled video into content and motion representations by explicitly modelling flow using image difference and use a single content encoding for future frame prediction. Similarly, MOCOGAN \cite{Tulyakov_2018_CVPR} disentangled video by sampling content representations once for the whole video sequence and motion representation for each frame, realism of the predicted frames is enforced using adversarial training. DRNET\cite{denton2017unsupervised} disentangled video into content and pose representations by using an adversarial loss term which aims to confuse a discriminator classifying pose vectors between same and different video sequences. DDPAE \cite{hsieh2018learning} decomposed the video into constituent set of objects and learn disentangled representations of content and pose for these objects. In contrast to these methods we learn to factorise video into content/pose by adopting the DRNET video generation architecture and penalising mutual information (MI) between pose latent variables across time.

\begin{figure*}
    \centering
    \includegraphics[width=0.90\textwidth]{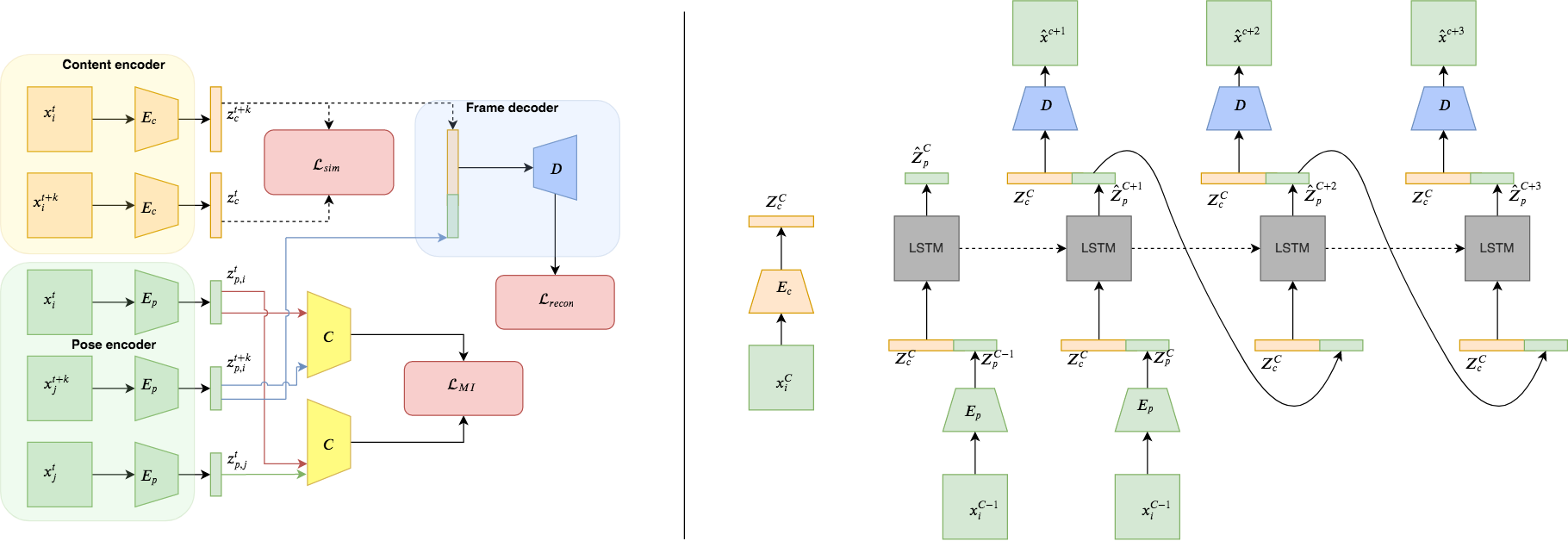}
    \caption{Left: Shows the training procedure for content encoder $E_c$, $E_p$ and $D$ along with various training objectives. To calculate MI, pose latent variables $z_{p,i}^{t}$, $z_{p,i}^{t+k}$ and $z_{p,j}^{t+k}$ are taken from sampled videos $x_i^{1:C+T}$ and $x_j^{1:C+T}$ by using pose encoder $E_p$, where $i$ and $j$ denote that they belong to two distinct video sequences. Joint samples $(z_{p,i}^t,z_{p,i}^{t+k})$ and marginal samples $(z_{p,i}^t,z_{p,j}^{t+k})$ are given as input to the critic $C$ and the outputs are used to estimate MI using equation \ref{eq:4}. Right: Shows the process of recurrent generation of pose latent variables $\hat{z}_p^t$ using LSTM network. These predicted pose vectors are used to generate future frames by decoder $D$.}
    \label{fig:my_label}
\end{figure*}

\subsection{Disentangling Representations}
Unsupervised learning of disentangled data representations is a well explored area in AI research \cite{bengio2013representation,lake2017building}. InfoGAN \cite{chen2016infogan} maximized MI between latent code and observations to disentangle the latent generative factors. Recent research has also used different variations of Evidence lower bound (ELBO) proposed by Kingama \cite{kingma2013auto} to enforce such disentanglement. $\beta$-VAE \cite{higgins2017beta} proposed a method based on using a high multiplicative constant for KL-divergence in ELBO to limit the capacity of latent information channel, thus enforcing highly factorised latent space representations. Total correlation based penalty between latent variables to attain disentanglement effect has also been explored by $\beta$-TCVAE \cite{chen2018isolating} and FactorVAE \cite{kim2018disentangling}. However, \cite{locatello2018challenging} theoretically explained that it is impossible to disentangle latent factors of data generation without exploiting the structure of the true generative factors. \cite{li2018disentangled,mathieu2016disentangling,hsu2017unsupervised,denton2017unsupervised} exploit such known structural relationships between factors of data generation to learn factorised latent representations. In our proposed MIPAE framework, video representations are factorized by considering the temporal structure of the content and pose generative factors, that is, the time independence of content and time dependence of pose.

\subsection{Mutual Information Estimation}
Mutual Information(MI) features prominently in ICA literature, such as \cite{roberts2001independent}, but they failed to provide a general method for computationally estimating mutual information from samples of two random variables. \cite{kraskov2004estimating}, \cite{ross2014mutual} proposed a non-parametric method to estimate mutual information between two random variables using k-nearest neighbour approach. But these methods do not scale well for large and high dimensional data.

Recent works have used variational estimation coupled with deep neural networks for tractable estimation of MI. \cite{nguyen2010estimating} proposed variational lower bound to estimate f-divergences, F-GAN \cite{nowozin2016f} utilised this to estimate common instances of f-divergence like KL-divergence or Jensen-Shannon divergence given samples from two different distributions. MINE \cite{belghazi2018mine} used Donsker-Vardhan \cite{donsker1983asymptotic} dual formulation for estimating MI, which is expressed as KL-divergence between joint distribution and the product of marginal distributions over a couple of random variables. Methods based on variational lower bounds use a deep neural network as critic to estimate MI. However, updating the critic using gradients of the F-GAN formulation is unstable. Hence, \cite{pmlr-v97-poole19a} proposed to use GAN objective instead of using gradients from F-GAN to train the critic, we use this estimate of MI in our work. For comprehensive overview on variational bounds of MI, readers are requested to refer to \cite{pmlr-v97-poole19a}. 

\section{Our Approach}

Our task is to generate the next $T$ frames, $\hat{x}^{C+1:C+T} = (\hat{x}^{C+1},\hat{x}^{C+2}\dots \hat{x}^{C+T})$ of a video sequence conditioned on a sub-sequence of $C$ context frames, $x^{1:C} = (x^1,x^2,\dots x^C)$. Making predictions in the high dimensional image space pose a major challenge to auto-regressive models. We propose some simplifying assumptions to overcome this challenge of learning a predictive distribution over possible futures in  image-pixel space. It is assumed that the identity of the constituent objects in the video sequence do not change with time, whereas the pose of these objects keep varying throughout the video sequence. Hence, the true generative factors for a video sequence can be considered to be time independent content factor, $f_c$, and time dependent pose factors, $f_p^{1:C+T}$. On top of the aforementioned assumption, we also consider that the video sequences are generated by a two step random process: (1) samples of the generative factors - content $f_c$ and pose $f_p^{1:C+T}$ are drawn from their true underlying joint distribution $P(f_c,f_p^{1:C+T})$; (2) after having obtained these factors, video frames are generated from the true conditional distribution $P(x^{1:C+T}|f_c,f_p^{1:C+T})$.

Based on these simplifying assumptions and two step video generative process, the latent space representation of video sequences are factorised into two factors, time independent content, $z_c$, and time dependent pose, $z_p^{1:C+T}$. This factorisation simplifies the task of making predictions in high dimensional pixel-space to predicting low dimensional pose representation $\hat{z}_p^{C+1:C+T}$ of the sequence. Future frames $\hat{x}^{C+1:C+T}$ are reconstructed from the predicted pose $\hat{z}_p^{C+1:C+T}$ and content $z_c$ representations. In the proposed framework, the time independent content factor $z_c$ is acquired by training a content encoder $E_c$ such that $z_c$ is acquired from a single frame $x^C$. Here $z_c$ is considered to be consistent for all the future frames. We also train a pose encoder $E_p$ which is applied to all frames independently to get the pose representations $z_p^{1:C+T}$. As the second step of the generative process under consideration, a decoder $D$ is trained to reconstruct $\hat{x}^t$ from $z_c$ and $z_p^t$, where $z_p^t$ denotes the pose representation for the frame $x^t$.

It is essential to note that in context of video prediction, proper pose/content disentanglement refers to a situation where any change in the true generative factors $f_c$ or $f_p^t$ should lead to change only in the learned content $z_c$ or pose $z_p^t$ representations, respectively. However, Locatello-Bauer \etal \cite{locatello2018challenging} showed that disentangling the generative factors of data in a completely unsupervised manner is impossible. This is because for any true generative factor $f$, one can always specify another equivalent set of generative factors $\hat{f}$ by trivially applying a deterministic function on $f$ such that $\hat{f}=g(f)$. Here, $\hat{f}$ is completely entangled with $f$. In unsupervised learning where only data samples are given, the learned latent representations $z$ cannot  distinguish the true generative factors $f$ from their transformed versions $\hat{f}$. Hence, it will be entangled with either $f$ or $\hat{f}$. In view of the above, it is proposed to disentangle video representations into pose $z_p^{1:C+T}$ and content $z_c$ latent representations by exploiting the temporal structure of the true underlying generative factors - time independence of content $f_c$ and time dependence of the pose $f_p^t$.

Formally, in MIPAE framework we enforce this temporal structure by applying similarity loss $\mathcal{L}_{sim}$ between the content latent representations $z_c$  of different frames from a given sequence and by minimizing the proposed Mutual Information Loss $\mathcal{L}_{MI}$ between the pose latent representations $z_p^t$ across time. $\mathcal{L}_{sim}$ in Fig. \ref{fig:my_label} (Left), is the $l_2$ loss minimised between encoded frames ${z_c}^t$, ${z_c}^{t+k}$ acquired by encoding frames ${x}^t$, ${x}^{t+k}$ by the content encoder $E_c$. $\mathcal{L}_{sim}$, enforces the content encoder $E_C$ to extract temporally invariant content features from the video sequence. However, this alone does not ensure proper pose/content disentanglement, as the pose latent representation of frames can still encode some aspect of the temporally invariant content information. Any content information in pose representations can be modeled as the mutual information between them. The pose encoder $E_p$ is restrained from encoding any content information by penalizing the mutual information between $z_p^t$ and $z_p^{t+k}$ for some random offset $k$.  As shown in Fig. \ref{fig:my_label} (Left), to calculate the mutual information between $z_p^t$ and $z_p^{t+k}$, we acquire the joint $(z_{p,i}^t,z_{p,i}^{t+k})$ and marginal samples $(z_{p,i}^t,z_{p,j}^{t+k})$, where $i$ and $j$ denote they belong to two distinct video sequences. A critic, $C$, is trained to discriminate between the joint and marginal samples. We use the critic outputs to estimate mutual information using $\mathcal{L}_{MI}$ as described in Eq. \ref{eq:4}. Application of $\mathcal{L}_{MI}$ is essential for proper pose/content factorisation of video representation, as the pose encoder $E_p$ and content encoder $E_c$ are explicitly constrained to encode complementary set of information for proper frame reconstruction. To ensure proper reconstruction, $l_2$ reconstruction error $\mathcal{L}_{recon}$ is minimised between the ground truth and decoded frame. Similarity loss $\mathcal{L}_{sim}$, reconstruction loss $\mathcal{L}_{recon}$ and proposed mutual information loss $\mathcal{L}_{MI}$ have been explained in section below.

\subsection{Models and Architecture}
The proposed architecture and loss terms are similar to DRNET except for their adversarial loss which is replaced with our mutual information objective. The different loss terms are explained below as follows:

\noindent\textbf{Similarity Loss:} To enforce time invariance of content representation we penalize change in content representation between two different frames from the same video sequence that are separated by random offset $k \in [0,K]$ time steps:
\begin{equation}\label{eq:2}
    \mathcal{L}_{sim} = \mathbb{E}_{P(x^t, x^{t+k})} \left[ \|E_c(x^t) - E_c(x^{t+k}) \|_2^2 \right]
\end{equation}

\begin{figure*}
    \centering
    \begin{tabular}{c|c|c}
    \subfloat[DRNET \cite{denton2017unsupervised}]{
        \begin{tabular}{cc}
            \\
            \\
            &\fcolorbox{green}{white}{\includegraphics[width=0.15\textwidth]{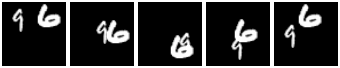}}\\
            \fcolorbox{red}{white}{\includegraphics[width=0.03\textwidth]{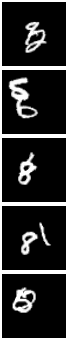}} & \includegraphics[width=0.15\textwidth]{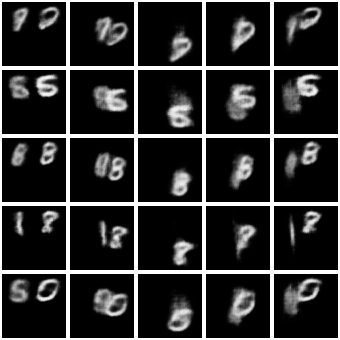}
        \end{tabular}\label{fig:mnist_drnet_swap}}&
    \subfloat[MIPAE]{
        \begin{tabular}{cc}
            \\
            \\
            &\fcolorbox{green}{white}{\includegraphics[width=0.15\textwidth]{mnist_source.png}}\\
            \fcolorbox{red}{white}{\includegraphics[width=0.03\textwidth]{mnist_content.png}} & \includegraphics[width=0.15\textwidth]{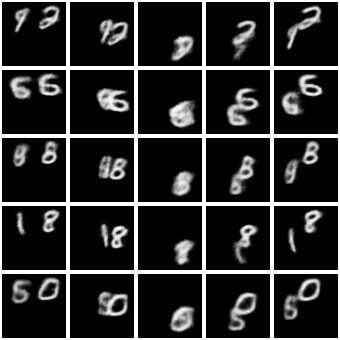}
        \end{tabular}\label{fig:mnist_ours_swap}}&
    \subfloat[Predictions]{
        \begin{tabular}{c}
            \includegraphics[width=0.4\textwidth]{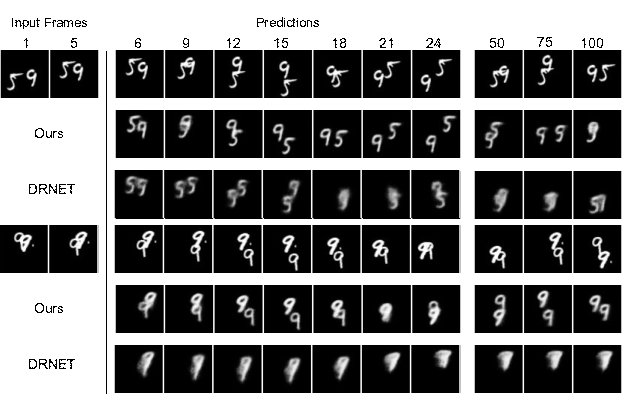}
        \end{tabular}\label{fig:mnist_pred}
    }\\
    \end{tabular}
    \caption{Qualitative comparison on moving MNIST dataset. Figures \protect\subref{fig:mnist_drnet_swap} and \protect\subref{fig:mnist_ours_swap} demonstrate pose-content disentanglement in DRNET and our model respectively. Each image in the grid is generated by taking pose latent variable from sequence highlighted in green and content latent variable from images highlighted in red. It can be seen that our model generates sharp frames in contrast to blurry predictions by DRNET in frames involving complex interactions between MNIST digits due to better content/pose disentanglement. Figure \protect\subref{fig:mnist_pred} show future frame predictions by our model and DRNET on two sequences. It is visually evident that our model produces future frames closer to ground truth frames over longer horizon.}
    \label{fig:qualitative_mnist}
\end{figure*}

\noindent\textbf{Mutual Information Loss:} We now introduce the proposed Mutual Information loss term which is applied on the pose latent variables across time to achieve proper pose/content factorisation. For estimating the mutual information between  $z_p^t$ and $z_p^{t+k}$, we first train a critic $C$ to classify whether $z_p^t$ and $z_p^{t+k}$ are sampled from joint distribution $P(z_p^t,z_p^{t+k})$ or the product of marginal distributions $P(z_p^t)P(z_p^{t+k})$ for some random offset $k \in [0,K]$ using standard GAN discriminator objective \cite{goodfellow2014generative}. In practice, samples of  $(z_p^t,z_p^{t+k})$ from $P(z_p^t,z_p^{t+k})$ are acquired by encoding frames $x^t$ and $x^{t+k}$ belonging to the same video sequence, whereas samples from the $P(z_p^t)P(z_p^{t+k})$ is acquired by encoding frames from two different video sequences, as shown in Fig. \ref{fig:my_label} (Left). The GAN discriminator objective is maximized for the optimal critic, $C^*(z_p^t,z_p^{t+k}) = log(P(z_p^t,z_p^{t+k})/P(z_p^t)P(z_p^{t+k}))$. The expectation of $C^*$ with respect to $P(z_p^t,z_p^{t+k})$ is the definition of MI. The critic objective is described below where $\sigma(.)$ is the sigmoid function which is applied on critic output:

\begin{equation}\label{eq:3}
    \begin{split}
        \mathcal{L}_{C} = \mathbb{E}_{P(x^t,x^{t+k})} \left[\sigma(C(E_p(x^t),E_p(x^{t+k})))\right] \\ +\mathbb{E}_{P(x^t)P(x^{t+k})} \left[1 - \sigma(C(E_p(x^t),E_p(x^{t+k})))\right]
    \end{split}
\end{equation}

 Unlike \cite{kim2018disentangling}, instead of estimating MI using monte-carlo estimation of the expectation of $C^*$ with respect to $P(z_p^t,z_p^{t+k})$ we use a variational lower bound of MI, $I_{JS}$ proposed in \cite{pmlr-v97-poole19a}. $I_{JS}$ exhibits lower variance in comparison to F-GAN and monte-carlo estimates:

\begin{equation}\label{eq:4}
    \begin{split}
        \mathcal{L}_{MI} = \mathbb{E}_{P(z_p^t,z_p^{t+k})} \left[C(z_p^t,z_p^{t+k})\right] \\
        -\mathbb{E}_{P(z_p^t)P(z_p^{t+k})} \left[\exp(C(z_p^t,z_p^{t+k}))\right]
    \end{split}
\end{equation}
By minimizing this MI estimate, $E_p$ is restricted from encoding any content information.

\noindent\textbf{Reconstruction Loss:} We use pixel-wise $l_2$ loss between decoded frame $D(E_c(x^t),E_p(x^t))$ and the ground truth frame $x^{t}$.
\begin{equation} \label{eq:1}
    \mathcal{L}_{recon} = \mathbb{E}_{P(x^t)} \left[ \|D(E_c(x^t),E_p(x^t)) - x^{t}\|_2^2\right]
\end{equation}

\noindent\textbf{Training Methodology:} The overall training objective for $E_c$, $E_p$ and $D$ is as follows:
\begin{equation}\label{eq:5}
    \mathop{min}_{E_c,E_p,D} \mathcal{L}_{recon} + \alpha \mathcal{L}_{sim} + \beta \mathcal{L}_{MI}
\end{equation}

Training object for the critic $C$ is given by:
\begin{equation}\label{eq:6}
    \mathop{max}_{C} \mathcal{L}_{C}
\end{equation}

\noindent\textbf{LSTM training procedure:} The LSTM $L$ is trained separately after training the main network, $E_c$, $E_p$ and $D$. To predict a future frame $\hat{x}^t$, first, the LSTM $L$ predicts $\hat{z}_p^{t}$ from previous frame's pose $\Tilde{z}_p^{t-1}$  and content representation $z_c^C$ of the last known frame $x^C$. Where $\Tilde{z}_p^{t-1}$ is the pose representation $z_p^{t-1}$ acquired from  $E_p$ if $t-1 \in [1,C]$, else $\Tilde{z}_p^{t-1}$ is $\hat{z}_p^{t-1}$, the pose representation predicted by $L$ for the frame $t-1$. 
\begin{equation}\label{eq:7}
    {\hat{z}_p}^t = L(z_c^C, \Tilde{z}_p^{t-1}) \ where\  \Tilde{z}_p^{t} = 
    \begin{cases}
    E_p(x^t) & t <  C + 1\\
    L(z_c^C, \Tilde{z}_p^{t-1}) & t \geq C + 1
    \end{cases}
\end{equation}   

The training objective for $L$ is to minimize the $l_2$ loss between predicted poses, $\hat{z}_p^{2:C+T}$, and  poses inferred from ground truth frames, $z_p^{2:C+T}$.
Decoder $D$ is used to generate the future frame $\hat{x}^t$ from the content $z_c$ and the predicted pose representation $\hat{z}_p^t$ of the future frame, such that $\hat{x}^t = D(z_c^C,\hat{z}_p^t)$. Note that content representation $z_c^C$ is fixed from the last known frame $x^C$, while pose representations are predicted in a recurrent manner. Fig. \ref{fig:my_label} (Right) depicts the future frame generation process. Model architecture for $E_p$, $E_c$, $C$, $L$ and $D$ are given in Section \ref{Traind}.

\subsection{Evaluation Metric}
A popular method to evaluate disentanglement is latent traversal where change in image reconstruction is observed against change in one dimension of the latent space by holding other dimensions constant.
This evaluation method is effective in finding methods that are unable to disentangle the generative factors of data but does not provide any quantitative measure of the effectiveness of disentanglement. Previous methods \cite{higgins2017beta,chen2018isolating,kim2018disentangling} propose various metrics to quantitatively evaluate the effectiveness of disentanglement for datasets with known generative factors. Methods in \cite{denton2017unsupervised,higgins2017beta} trained a classifier to predict ground truth factors from latent variables and use the classification accuracy as an indicator of effective disentanglement. As pointed out by \cite{chen2018isolating}, these  methods fail to detect cases where latent representations and true factors are not axis-aligned. These methods also depend extensively on weight initialisation and training hyperparameters. H. Kim \etal \cite{kim2018disentangling} used a majority vote classifier to avoid the dependency on hyperparameters and also deal with the axially unaligned case. However, these classification based evaluation techniques generally need further improvements.

T. Q. Chen \etal \cite{chen2018isolating} proposed a generic metric Mutual Information Gap metric (MIG). In this formulation, mutual information is calculated between each pair of dimensions of the true factors and learned representations. MIG can be used in scenarios where mutual information can be calculated (i.e where factors of data generation are known a priori). The generic MIG metric is modified to evaluate the effectiveness of disentanglement for video prediction methods. In our adaptation of the MIG metric for video prediction, mutual information is calculated between generative factors and the learned pose, content representations, instead of independent dimensions. We use this adopted version of MIG metric to quantitatively evaluate the effectiveness of disentanglement between the learned pose and content latent representations. Any subsequent reference to the mutual information gap (MIG) metric refers to the modified version described below:   

\begin{equation}\label{eq:8}
\begin{split}
    MIG = \frac{0.5}{H(f_c)} \Big(I(f_c,z_c) - I(f_c,z_p)\Big) \\
    + \frac{0.5}{H(f_p)} \Big(I(f_p,z_p) - I(f_p,z_c) \Big)
\end{split}
\end{equation}

\begin{figure*}
    \centering
    \begin{tabular}{c|c|c}
    \subfloat[DRNET\cite{denton2017unsupervised}]{
        \begin{tabular}{cc}
            \\
            \\
            &\fcolorbox{green}{white}{\includegraphics[width=0.15\textwidth]{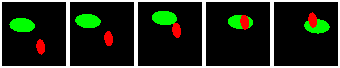}}\\
            \fcolorbox{red}{white}{\includegraphics[width=0.03\textwidth]{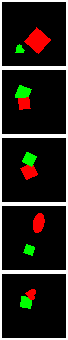}} & \includegraphics[width=0.15\textwidth]{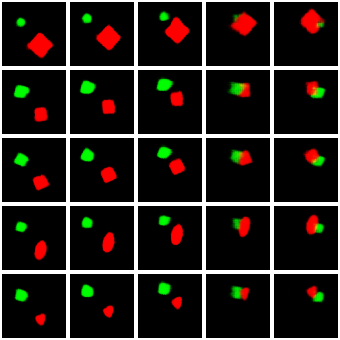}
        \end{tabular}\label{fig:dsprites_drnet_swap}}&
    \subfloat[MIPAE]{
        \begin{tabular}{cc}
            \\
            \\
            &\fcolorbox{green}{white}{\includegraphics[width=0.15\textwidth]{dsprites_source.png}}\\
            \fcolorbox{red}{white}{\includegraphics[width=0.03\textwidth]{dsprites_content.png}} & \includegraphics[width=0.15\textwidth]{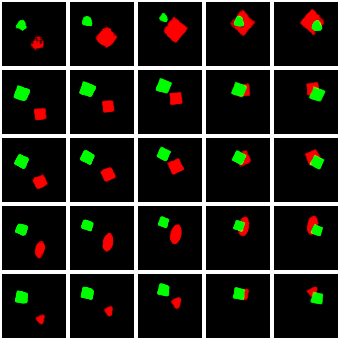}
        \end{tabular}\label{fig:dsprites_ours_swap}}&
    \subfloat[Prediction]{
        \begin{tabular}{c}
            \includegraphics[width=0.4\textwidth]{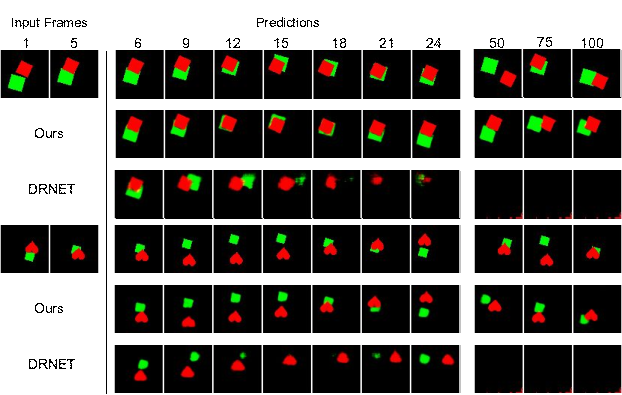}
        \end{tabular}\label{fig:dsprites_pred}
    }\\
    \end{tabular}
    \caption{Qualitative comparison of disentanglement on moving DSprites. Figures \protect\subref{fig:dsprites_drnet_swap} and \protect\subref{fig:dsprites_ours_swap} demonstrate pose-content disentanglement in DRNET and our model respectively. Each image in the grid is generated by taking pose latent variable from sequence highlighted in green and content latent variable from images highlighted in red. It can be seen that DRNET is unable to produce accurate shape of different objects , where as our model produces sharp reconstructions capturing true object shapes due to proper pose/content factorisation. Figure \protect\subref{fig:dsprites_pred} shows future frames generated by our method and DRNET on moving DSprites dataset. It can be seen that in comparison to DRNET, MIPAE generates coherent and stable long range predictions.}
    \label{fig:qualitative_dsprites}
\end{figure*}

In the above formulation of MIG, $H(.)$ refers the entropy of a random variable and $I(.,.)$ denotes mutual information between two random variables. Any model that fails to learn proper pose/content disentanglement of video representation will exhibit low score in the proposed MIG metric. The first term will have small value when pose latent variables partially encode content descriptions. Similarly, if content latent variable encodes pose descriptions, the second additive term will have small value. In either case of improper pose/content disentanglement, the MIG score will exhibit low score.
In our experiments, true generative factors $f_c$ and $f_p$ are discrete random variables whereas learned representations $z_c$ and $z_p$ are continuous random variable, so we use \cite{ross2014mutual}  to estimate the mutual information between them, while \cite{grassberger2003entropy} is used to calculate entropy of the discrete generative factors.   
\begin{figure*}
    \centering
    \begin{tabular}{ccc}
        \includegraphics[trim=35 0 50 35,clip,width=0.3\textwidth]{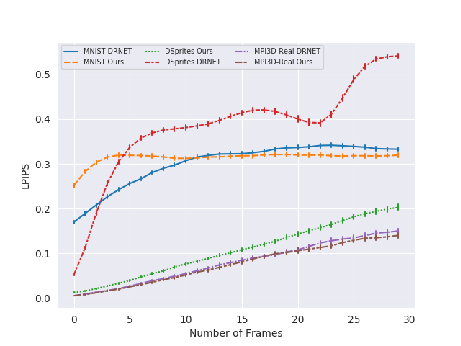}&
        \includegraphics[trim=35 0 50 35,clip,width=0.3\textwidth]{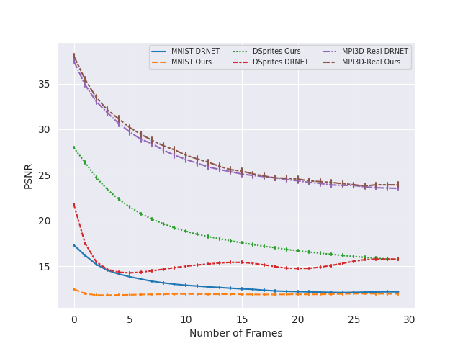}&
        \includegraphics[trim=35 0 50 35,clip,width=0.3\textwidth]{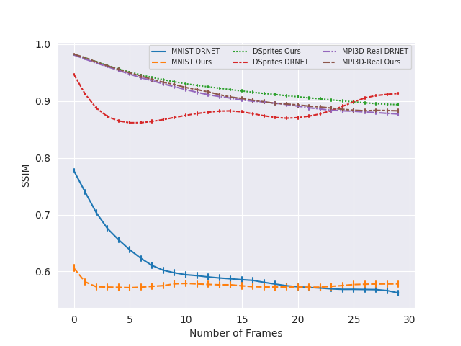}
    \end{tabular}
    \caption{Quantitative comparison on SM-MNIST, Dsprites and MPI3D-Real datasets of future frame prediction over long range. Left graph shows the LPIPS distance (lower is better) between ground truth frames and predicted frames. Middle and Right graphs show the PSNR and SSIM metric (higher is better) between ground truth and predicted frames.}
    \label{fig:quantitative}
\end{figure*}
\begin{table}[t]
    \setlength{\tabcolsep}{0.8pt}
    \renewcommand{\arraystretch}{1.3}
    \caption{MIG metric details}
    \label{mig_table}
    \centering
    \begin{tabular}{|cc|c|c|c|c|c|}
        \hline
        Dataset & Experiment & $I(f_c,z_c)$ & $I(f_c,z_p)$ & $I(f_p,z_c)$ & $I(f_p,z_p)$ & $MIG$\\
        \hline
        \hline
        \multirow{2}{*}{Dsprites} & DRNET\cite{denton2017unsupervised} & 5.6476 & 0.7483 & 0.0748 & 6.3434 & 0.8574 \\
        &Ours & 5.6992 & 0.4660 & 0.725 & 6.4977 & \textbf{0.8975} \\
        \hline
        \multirow{2}{*}{MPI3D Real} & DRNET\cite{denton2017unsupervised} & 8.1353 & 0.0376 & 0.0448 & 6.2029 & 0.5658 \\
        & Ours & 8.3866 & 0.0461 & 0.0080 & 7.1034 & \textbf{0.6126}\\
        \hline
    \end{tabular}
\end{table}

\section{Experiments}
We evaluate our MIPAE framework quantitatively and qualitatively on two synthetic datasets, moving MNIST, Dspites \cite{dsprites17}, and one real world moving MPI3D-Real \cite{gondal2019transfer} dataset. We use the LPIPS distance\cite{zhang2018perceptual} measure between predicted and ground truth frames to substantiate the visual superiority of predicted frames by our method on all three datasets. Previous works have evaluated the fidelity of their frame predictions by using PSNR and SSIM scores. PSNR and SSIM metric based evaluations show poor correspondence with the visual quality of the predicted frames \cite{lee2018stochastic}. However, for the purpose of completeness, PSNR and SSIM scores have also been calculated for comparative evaluation f0of MIPAE. We provide quantitative comparison on the effectiveness of disentanglement achieved by MIPAE with DRNET by using the proposed $MIG$ score on datasets with known generative factors namely,  DSprites and MPI3D-Real.

Code for MIPAE and the experiments are available at \url{https://cvit.iiit.ac.in/projects/mutualInfo/}.

\subsection{Training Details} \label{Traind}
For moving MNIST and DSprites datasets, $E_c$, and $E_p$ all use DCGAN \cite{dcgan} architecture with $\|z_c\|=128$ and $\|z_p^t\|=5$. $D$ is the mirrored version of the encoder where sub-sampling convolutional layers are replaced with deconvolutional layers.

For moving MPI3D-Real, $E_p$ is ResNet-18 \cite{he2016deep} architecture and $E_c$ and $D$ are VGG-16 \cite{simonyan2014deep} architecture with $\|z_c\|=128$ and $\|z_p^t\|=10$. Decoder $D$ is the mirrored version of the content encoder $E_c$. In the decoder network, spatial up-sampling layers are used instead of pooling layers. We also provide skip-connection from content encoder to decoder in U-Net \cite{ronneberger2015unet} style architecture.

In all experiments, the critic $C$ is a multi-layer perceptron with two hidden layers of 512 units each and RELU activation function. We use Adam optimizer \cite{kingma2014adam} with learning rate 0.002 and $\beta_1=0.5$. We choose $\alpha=1$ and $\beta=0.0001$ for our training objective as described in equation \ref{eq:5}.

Recurrent pose prediction network $L$ is a two layer LSTM network with 256 cells each, with linear input and output embedding layers. The proposed MIPAE is trained to observe 5 context frames to predict 10 future frames. For fair comparison, both MIPAE and DRNET are trained for the same number of video frames and hyper-parameters. 

\subsection{Moving MNIST}
Moving MNIST dataset consists of two MNIST digits bouncing independently in a 64x64 image. MNIST digits move with constant velocity and undergo mirror reflection upon collision with frame boundaries. This dataset has been widely used by previous works \cite{denton2017unsupervised,hsieh2018learning} to substantiate  their disentanglement claims. Fig. \ref{fig:qualitative_mnist}\subref{fig:mnist_drnet_swap} and Fig. \ref{fig:qualitative_mnist}\subref{fig:mnist_ours_swap} qualitatively demonstrate the pose/content disentanglement of DRNET and our model respectively. In these figures new frames in the grid are generated by taking position latent variables $z_p^{1:C+T}$ from the sequence highlighted in green and content latent variables $z_c$ from the images highlighted in red. In our predictions, digits in the generated sequences imitate the pose of digits in the source sequence (highlighted in green) irrespective of the input content. This indicates that that our pose representations are truly content agnostic. As mentioned in \cite{denton2017unsupervised}, DRNET requires additional information in the form of colored digits to learn disentangled representations. Fig. \ref{fig:qualitative_mnist} shows that DRNET produces blurry results when trained without color, in contrast to our method which produces sharp digits even without color information. This is indicative of the fact that our model attains a better pose/content disentanglement even without additional color coding of the MNIST digits.

Fig. \ref{fig:qualitative_mnist}\subref{fig:mnist_pred} shows frame prediction by our model and DRNET on two sequences. Our model produces predictions closer to ground truth for longer horizons, as depicted in the first predicted sequence, indicating that higher pose/content disentanglement helps in sustained long range prediction. This can be verified quantitatively in Fig. \ref{fig:quantitative}, that frames predicted by our model has lower LPIPS distance with ground truth frames over longer horizon. Further, in Fig. \ref{fig:qualitative_mnist}\subref{fig:mnist_pred} it is noteworthy that our model is able to generate distinct digits in video sequences containing multiple instances of the same digit, whereas DRNET is unable to keep track and confuses between these mulitple instances due to lack of colour information in the content representaion of MNIST digits. We are unable to evaluate the effectiveness of disentanglement using the proposed $MIG$ metric due to partial knowledge of the generative factors for MNIST videos, specifically, the true content generative factors $f_c$ of MNIST digits are unknown.

\subsection{Moving DSprites}
DSprites \cite{dsprites17} is a procedurally generated dataset  with known generative factors. The dataset contains 3 different shapes, at 6 scales and 40 rotational orientations. Video sequences form this dataset are generated in a controlled manner by moving the shapes with constant velocity which undergo mirror reflection upon collision with frame boundaries. Fig. \ref{fig:qualitative_dsprites}\subref{fig:dsprites_drnet_swap} and Fig. \ref{fig:qualitative_dsprites}\subref{fig:dsprites_ours_swap} shows qualitative disentanglement results of DRNET and our model respectively. DRNET is unable to produce accurate reconstruction of the shapes, where as our method produces sharp and accurate reconstruction of different objects in the frames. This shows that our method is able to capture better content representations $z_c$ due to effective pose/content disentanglement. Our disentanglement claim is validated by the proposed MIG metric as the true generative factors for this dataset are known apriori. MIG metric score of our method and DRNET can be found in Tab.\ref{mig_table} along with the estimated MI between generative factors and learned representations. Our method has a higher MIG score as compared to DRNET indicating better pose/content disentanglement. This finding is further supported by visual comparison of generated future frames by both methods, as depicted in Fig. \ref{fig:qualitative_dsprites}\subref{fig:dsprites_pred}, it can be seen that DRNET is unable to sustain frame prediction over longer horizons. MIPAE outperforms DRNET on PSNR, SSIM and LPIPS metrics, shown in Fig. \ref{fig:quantitative}.

\subsection{Moving MPI3D-Real}
\begin{figure}
    \setlength{\textfloatsep}{-10pt}
    \centering
    \subfloat[DRNET\cite{denton2017unsupervised}]{
      \begin{tikzpicture}[inner sep=0pt, spy using outlines = {rectangle,red,magnification=2.5,size=0.75cm,connect spies}]
        \node (content) [rectangle,draw=red,line width = 2pt,anchor=south west] {\includegraphics[width=0.05\textwidth]{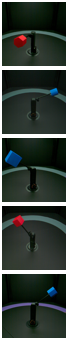}};
        \node (recon) [anchor=west] at ($(content.east) + (0.5pt,0)$) {\includegraphics[width=0.25\textwidth]{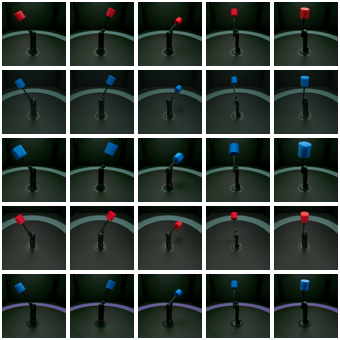}};
        \node (source) [rectangle,draw=green,line width = 2pt,anchor=south] at ($(recon.north)+(0,0.5pt)$) {\includegraphics[width=0.25\textwidth]{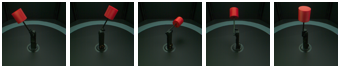}};
        \spy on (5.05,2.55) in node (zoom1) [anchor=west] at ($(recon.east) + (0.3cm,0)$);
        \spy on (5.05,1.65) in node (zoom2) [anchor=north] at ($(zoom1.south) + (0,-0.1cm)$);
      \end{tikzpicture}
      \label{fig:mpi_drnet_swap}}\qquad
    \subfloat[MIPAE]{
        \begin{tikzpicture}[inner sep=0pt, spy using outlines = {rectangle,red,magnification=2.5,size=0.75cm,connect spies}]
        \node (content) [rectangle,draw=red,line width = 2pt,anchor=south west] {\includegraphics[width=0.05\textwidth]{mpi_content.png}};
        \node (recon) [anchor=west] at ($(content.east) + (0.5pt,0)$) {\includegraphics[width=0.25\textwidth]{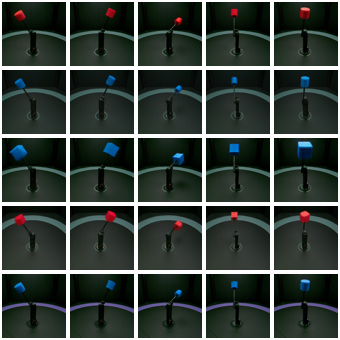}};
        \node (source) [rectangle,draw=green,line width = 2pt,anchor=south] at ($(recon.north)+(0,0.5pt)$) {\includegraphics[width=0.25\textwidth]{mpi_source.png}};
        \spy on (5.05,2.55) in node (zoom1) [anchor=west] at ($(recon.east) + (0.25cm,0)$);
        \spy on (5.05,1.65) in node (zoom2) [anchor=north] at ($(zoom1.south) + (0,-0.1cm)$);
      \end{tikzpicture}\label{fig:mpi_ours_swap}}
        \caption{Qualitative comparison of disentanglement on MPI 3D Real. Figures \protect\subref{fig:mpi_drnet_swap} and \protect\subref{fig:mpi_ours_swap} demonstrate pose-content disentanglement in DRNET and our model respectively. Each image in the grid is generated by taking pose latent variable from sequence highlighted in green and content latent variable from images highlighted in red. It can be seen that DRNET reconstructs cube as cylinder (the magnified part) where as our method reconstructs correctly.}
        \label{fig:qualitative_mpi}
\end{figure}

MPI3D-Real \cite{gondal2019transfer} is a real world dataset with known generative factors. It contains images of different objects mounted on a rotating robotic arm at different angular positions. Video sequences are generated by rotating the robotic arm with constant angular velocity which undergoes mirror reflection upon collision with ground. Fig. \ref{fig:qualitative_mpi}\subref{fig:mpi_drnet_swap} and Fig. \ref{fig:qualitative_mpi}\subref{fig:mpi_ours_swap} show the qualitative disentanglement results of DRNET and our method respectively. It can be seen that DRNET is unable to reconstruct accurate object shape, specifically in the third and fourth objects from the top where it reconstructs a cube as a cylinder, where as our method is able to produce sharp reconstruction of the cube. The quantitative comparison of future frames predicted can be found in Fig. \ref{fig:quantitative}. $MIG$ scores in Tab.\ref{mig_table} also indicate that our method has attains better disentanglement than DRNET.
\section{Conclusion and Future Work}
In this work, we propose MIPAE framework, that reduces the task of predicting high dimensional future frames by disentangling video representations into content and low dimensional pose latent factors which are easier to predict. This pose/content factorisation is achieved by penalising mutual information between pose latent variables across time. We also adopt Mutual Information Gap (MIG) metric to quantitatively compare  the effectiveness of disentanglement of the proposed method with DRNET to conclusively demonstrate that our method learns substantially better disentangled latent representations, which in turn leads to visually sharp and realistic frame prediction. These improvement over DRNET are achieved by replacing their adversarial loss with our mutual information loss without making any significant training procedure or architectural changes. The simplicity of this mutual information loss based disentanglement approach lends itself to easy integration with other video prediction models. In future we would like to extend our approach to VAE-based stochastic video prediction models.
{\small
\bibliographystyle{IEEEtran}
\bibliography{root}
}

\end{document}